\documentclass[twoside]{article}

\usepackage[accepted]{aistats2020}

\usepackage[round]{natbib}

\usepackage{amsmath,amssymb,amsfonts}
\usepackage{mathtools}
\usepackage{algorithm}
\usepackage{algorithmicx}
\usepackage{algpseudocode}
\usepackage{graphicx}
\usepackage{textcomp}
\usepackage{hyperref}
\usepackage{xcolor}
\def\BibTeX{{\rm B\kern-.05em{\sc i\kern-.025em b}\kern-.08em
    T\kern-.1667em\lower.7ex\hbox{E}\kern-.125emX}}
\usepackage{balance}

\usepackage{amsthm}
\usepackage{bm}
\usepackage{dsfont}

\newcommand{\defeq}{\triangleq}

\DeclareMathOperator*{\argmin}{\mathop{\arg\min}}
\DeclarePairedDelimiter\floor{\lfloor}{\rfloor}

\renewcommand{\vec}[1]{\bm{\mathrm{#1}}}
\newcommand{\x}{\vec{x}}
\newcommand{\y}{\vec{y}}
\newcommand{\V}{\vec{V}}
\renewcommand{\S}{\mathcal{S}}
\newcommand{\R}{\mathbb{R}}
\newcommand{\EE}[1]{\mathbb{E} \left[ #1 \right]}
\renewcommand{\L}{\vec{L}}
\newcommand{\crit}{\mathrm{crit}}
\newcommand{\Thetatilde}{\widetilde{\Theta}}

\newtheorem{theorem}{Theorem}
\numberwithin{theorem}{section}
\newtheorem{lemma}[theorem]{Lemma}
\newtheorem{proposition}[theorem]{Proposition}
\theoremstyle{definition}
\newtheorem{definition}[theorem]{Definition}
\theoremstyle{remark}
\newtheorem{remark}[theorem]{Remark}

\newcommand\algoname{GrAPL}

\begin{document}

\twocolumn[

\aistatstitle{Thresholding Graph Bandits with \algoname{}}

\aistatsauthor{ Daniel LeJeune \And Gautam Dasarathy \And Richard G. Baraniuk }

\aistatsaddress{ Rice University \And Arizona State University \And Rice University } ]

\begin{abstract}
In this paper, we introduce a new online decision making paradigm that we call \emph{Thresholding Graph Bandits}. The main goal is to efficiently identify a subset of arms in a multi-armed bandit problem whose means are above a specified threshold. While traditionally in such problems, the arms are assumed to be independent, in our paradigm  we further suppose that we have access to the similarity between the arms in the form of a graph, allowing us to gain information about the arm means with fewer samples. Such a feature is particularly relevant in modern decision making problems,  where rapid decisions need to be made in spite of the large number of options available. We present \algoname{}, a novel algorithm for the thresholding graph bandit problem. We demonstrate theoretically that this algorithm is effective in taking advantage of the graph structure when the structure is reflective of the distribution of the rewards. 
We confirm these theoretical findings via experiments on both synthetic and real data.  
\end{abstract}

\section{INTRODUCTION}
\label{sec:intro}

Systems that recommend products, services, or other attention-targets have become indispensable in the effective curation of information. Such personalization and recommendation techniques have become ubiquitous not only in product/content recommendation and ad placements  but also in a wide range of applications like drug testing, spatial sampling, environmental monitoring, and rate adaptation in communication networks; see, e.g., \citet{villar2015multi, combes2014optimal, srinivas2009gaussian}. These are often modeled as sequential decision making or \emph{bandit problems}, where an algorithm needs to choose among a set of decisions (or arms) sequentially to maximize a desired performance criterion. 

Recently, an important variant of the bandit problem was proposed by \citet{locatelli2016optimal} and \citet{gotovos2013active}, where the goal is to rapidly identify all arms that are above (and below) a fixed threshold. This \emph{thresholding bandit} framework, which may be thought of as a version of the combinatorial pure exploration problem \citep{chen2014combinatorial}, is useful in various applications like environmental monitoring, where one might want to identify the hypoxic (low-oxygen-content) regions in a lake;  like crowd-sourcing, where one might want to keep all workers whose productivity trumps the cost to hire them; or like political polling, where one wants to identify which political candidate individual voting districts prefer. Such a procedure may even be considered in human-in-the-loop machine learning pipelines, where the algorithm might want to select a set of options that meet a certain cut-off for closer examination by a human expert. 

In many important applications, however, one is faced with an enormous number of arms that need to sorted through almost instantaneously. This makes prior approaches untenable both from a computational and from a statistical viewpoint. However, when there is {\em information sharing} between these arms, one might hope that this situation can be improved.

In this paper, we consider the thresholding bandit problem in the setting where a graph describing the similarities between the arms is available (see Section~\ref{sec:setup-algo}). We show that if one leverages this graph information, and more importantly the \emph{homophily} (that is, that strong connection implies similar behavior), then one can achieve significant gains over prior approaches. We develop a novel algorithm, \algoname{} (see Section~\ref{sec:grapl}), that explicitly takes advantage of the graph structure and the homophily. We then characterize, using rigorous theoretical estimates of the error of \algoname{}, how this algorithm indeed leverages this side information to improve upon prior algorithms in similar settings. Finally, in Section~\ref{sec:experiments}, we confirm these theoretical findings via experiments on real and synthetic data.  

\section{THRESHOLDING GRAPH BANDITS}
\label{sec:setup-algo}
\subsection{Thresholding Bandits}

Let $N$ denote the number of bandit arms, which are observable via independent samples of the corresponding \emph{R-sub-Gaussian} distributions $\nu_i, \; i \in [N]$. That is, each distribution $\nu_i$ satisfies the following condition for all $t \in \R$:
\begin{align}
    {\mathbb{E}_{X \sim \nu_i} \left[\exp\{t (X - \mu_i)\} \right] \leq \exp\{R^2 t^2 / 2\}},
\end{align}
where $\mu_i = \mathbb{E}_{X \sim \nu_i}[X]$. The goal of a learning algorithm in the thresholding bandit problem is to recover the superlevel set $\S_\tau = \{i : \mu_i \geq \tau \}$ from these noisy observations. The learning algorithm is allowed to run for $T$ iterations, and at each iteration $t \in [T]$ it can select one arm $\pi_t \in [N]$ from which to receive an observation. At the end of the $T$ iterations, the algorithm returns its estimate $\widehat{\S}$ of the superlevel set $\S_\tau$. This variant of the multi-armed bandit problem was introduced by \citet{locatelli2016optimal}, who provided the Anytime Parameter-free Thresholding (APT) algorithm for solving the problem with matching upper and lower bounds. \citet{mukherjee2017thresholding} and \citet{zhong2017asynchronous} have since provided algorithmic extensions to APT that incorporate variance estimates and provide guarantees in asynchronous settings. Recently, \citet{tao2019thresholding} introduced the Logarithmic-Sample Algorithm and proved it to be instance-wise asymptotically optimal for minimizing aggregate regret.

The thresholding bandit problem can be thought of as a version of the combinatorial pure exploration (CPE) bandit problem described by \citet{chen2014combinatorial}.
As such, the appropriate performance loss measures the quality of the returned superlevel set estimate $\widehat{\S}$ at time $T$ rather than a traditional notion of regret. We adopt a natural loss function for this setting (as done by \citet{locatelli2016optimal}):
\begin{align}
    \mathcal{L}_T = \mathds{1} \left\{ \left| (\S_{\tau + \varepsilon} \cap \widehat{\S}^c) \cup (\S_{\tau - \varepsilon}^c \cap \widehat{\S}) \right| > 0  \right\},
\end{align}
which for any $\varepsilon > 0$ is the indicator that at least one $i$ such that $|\mu_i - \tau| > \varepsilon$ has been classified as being on the wrong side of the threshold. 

Next, we need a notion of complexity that captures the statistical difficulty of performing the thresholding. Towards this end, we set $\Delta_i \defeq \Delta_i^{\tau, \varepsilon} = |\mu_i - \tau| + \varepsilon$, where $\varepsilon$ is the same quantity as in the definition of $\mathcal{L}_T$, and define the \emph{complexity} of the thresholding problem as
\begin{align}
    H \defeq H_{\tau, \varepsilon} = \sum_{i=1}^N \Delta_i^{-2}.
\end{align}
This definition of complexity also plays a key role in the analysis of \citet{locatelli2016optimal}. Intuitively, if there are values $\mu_i$ that are near the threshold, then the superlevel set will be ``hard'' to identify, and the problem complexity $H$ will be correspondingly high. Conversely, if the values $\mu_i$ are far from the threshold, then the superlevel set will be ``easy'' to identify, and the problem complexity is correspondingly small.

\subsection{Thresholding Graph Bandits}

As discussed in the introduction, the main contribution of this paper is to present a new framework for such thresholding bandit problems where one has access to additional information about the similarities of arms. In particular, we will model this additional information as a weighted graph that describes the arm similarities. 
Let $\mathcal{G} = (\mathcal{V}, \mathcal{E}, \vec{W})$ denote a similarity graph defined on the $N$ arms such that each arm is a vertex in $\mathcal{V}$ and $\vec{W}\in \mathbb{R}^{N\times N}$ describes the weights of the edges $\mathcal{E}$ between these vertices. Let ${\L = \vec{D} - \vec{W}}$ denote the graph Laplacian, where $\vec{D} = \mathrm{diag}(\vec{W} \vec{1})$ is a diagonal matrix containing the weighted degrees of each vertex. The graph Laplacian in this context is functionally quite similar to the precision matrix of a Gaussian graphical model defined on the same graph,
where edges on the graph indicate conditional dependencies between two arms given all other arms, and the weight indicates the strength of the partial correlation.

The main idea behind leveraging this similarity graph is that, if the learning algorithm is aware of the similarity structure among arms through the graph $\mathcal{G}$, and if the rewards $\vec{\mu} = (\mu_i)_{i=1}^N$ vary smoothly among similar arms, then the learning algorithm can leverage the information sharing to avoid oversampling similar arms. 

We capture the effectiveness of the graph in helping with the information sharing using two related notions of complexity. The first is $\|\vec{\mu}\|_{\L_\lambda} = \sqrt{\vec{\mu}^\top \L_\lambda \vec{\mu}}$, the  $\L_\lambda$ norm of $\vec{\mu}$, where $\L_\lambda = \L + \lambda \mathbf{I}$ for some $\lambda>0$. It is not hard to check that this value is smaller for those $\vec{\mu}$'s that are smooth on the graph $\mathcal{G}$ (see, e.g., \citet{ando2007learning}). The second notion of complexity, the \emph{effective dimension}, characterizes the helpfulness of the graph itself.

\begin{definition}[{\citealp[Def.~1]{valko2014spectral}}]
\label{def:effective-dimension}
For any $\gamma>0$, $T\in \{1,2,\ldots,N\}$, the \textbf{effective dimension} $d_T$ of the regularized Laplacian $\L_{\lambda}$ is the largest
$d$ such that
\begin{equation}
    (d - 1) \gamma \lambda_d \leq \frac{T}{\log(1 + T/\gamma \lambda)},\label{eq:eff-dim}
\end{equation}
where $\lambda_d$ is the $d$-th eigenvalue of $\L_\lambda$ when $\lambda_1 \leq \ldots \leq \lambda_N$. 
\end{definition}
In Definition~\ref{def:effective-dimension}, $T$ is the time horizon of the algorithm; if $T>N$, then one may use $N$ instead of $T$ on the right side of \eqref{eq:eff-dim}.
$\gamma > 0$ is a free parameter that can be tuned in the algorithm design (see Section~\ref{sec:grapl}).

It can be checked  readily that the effective dimension is no larger than $N$ for any graph. In fact, as observed by \citet{valko2014spectral}, for many graphs of interest the effective dimension turns out to be significantly smaller than $N$. As we will see in Section~\ref{sec:grapl}, this quantity plays a key role in capturing the effectiveness of our algorithm in leveraging the arm-similarity graph.\footnote{We also note here that the same authors proposed an improved definition of effective dimension that is even smaller and remains applicable in our setting (see Section 1.3.1 of \citet{valko2016bandits}).}

\subsection{A Non-adaptive Approach}

Before introducing our algorithm for thresholding graph bandits, we first introduce a useful baseline.

Our algorithm for thresholding graph bandits has two primary components. The first of these is using the graph structure to regularize the estimate of the arm means using Laplacian regularization techniques, which have received considerable attention in recent decades \citep[see][]{belkin2005manifold, zhu2003semi, ando2007learning}. The second is an adaptive sampling strategy in the style of the Anytime Parameter-free Thresholding (APT) algorithm of \citet{locatelli2016optimal}.
In this section, we describe an algorithm which has only the first component---i.e., an algorithm which uses graph-regularized estimates of the arm means but selects which arm to sample next non-adaptively; see \mbox{Algorithm}~\ref{alg:nonadaptive}.

\begin{algorithm}
\caption{Thresholding via non-adaptive graph-regularized estimation}
\label{alg:nonadaptive}
\begin{algorithmic}[1]
\State{\textbf{Input:} $\tau, \varepsilon, \L, \gamma, T$}
\State{$\V_0 \gets \L + \lambda \vec{I}$}
\State{$\widehat{\vec{\mu}}_0 \gets \tau \vec{1}$}
\State{$\vec{n}_0 \gets \vec{0}$}
\For{$t$ in $1, \ldots, T$}
    \State{Determine $\pi_t$ non-adaptively}
    \State{Observe $x_t \sim \nu_{\pi_t}$}
    \State{$\V_t \gets \V_{t-1} + \gamma^{-1} \vec{e}_{\pi_t} \vec{e}_{\pi_t}^\top $}
    \State{$\x_t \gets \x_{t-1} +  \gamma^{-1} x_t \vec{e}_{\pi_t}$}
    \State{$\widehat{\vec{\mu}}_t \gets \V_t^{-1} \x_t$}
\EndFor
\State{\textbf{Output:} $\widehat{\S} = \left\{ i : \widehat{\mu}_i^T \geq \tau \right\}$}
\end{algorithmic}
\end{algorithm}

At each iteration, Algorithm~\ref{alg:nonadaptive} first selects an arm to sample in a non-adaptive manner. This could be simply the selection of an arm at random or cycling through a permutation of the arms, for example.

Next, the algorithm solves the following Laplacian-regularized least-squares optimization problem for some $\gamma > 0$:
\begin{align}
    \widehat{\vec{\mu}}_t = \argmin_{\vec{\mu}} \sum_{s=1}^t (x_s - \mu_{\pi_s})^2 + \gamma \|\vec{\mu}\|_{\L_\lambda}^2.
\end{align}
This optimization problem is known to promote solutions that are \emph{smooth} across the graph \citep[see][]{ando2007learning}. In fact, let $\vec{e}_i$ denote the $i$-th standard basis vector and recall that $\pi_t$ denotes the index of the arm pulled at time $t$. If we define the quantities
\begin{align}
    \V_t &= \L_\lambda + \frac{1}{\gamma}\sum_{s=1}^t \vec{e}_{\pi_s} \vec{e}_{\pi_s}^\top , \\
    \label{eq:x}
    \x_t &= \frac{1}{\gamma}\sum_{s=1}^t x_s \vec{e}_{\pi_s},
\end{align}
then the above optimization problem admits a solution of the form
\begin{equation}
    \label{eq:muhat}
    \widehat{\vec{\mu}}_t = \V_t^{-1} \x_t.
\end{equation}
We note that this solution also corresponds to \emph{a posteriori} estimation of $\vec{\mu}$ under a Gaussian prior with precision matrix $\L_\lambda$ when the distributions $\nu_i$ are Gaussian with variance $R^2$. The following proposition characterizes the performance of Algorithm~\ref{alg:nonadaptive}.

\begin{proposition}
    If Algorithm~\ref{alg:nonadaptive} is run using a sampling strategy where every $N$ iterations all arms are sampled, and $\|\vec{\mu}\|_{\L_\lambda} \leq \sqrt{\frac{T}{\gamma \widetilde{H}}}$, then for $T = kN$ for any positive integer $k$,
    \begin{align}
        \label{eq:nonadaptive-ub}
        \EE{\mathcal{L}_T} \leq \exp \Bigg\{ &- \frac{\gamma^2}{2R^2} \left( \sqrt{\frac{T}{\gamma \widetilde{H}}} - \|\vec{\mu}\|_{\L_{\lambda}} \right)^2 \nonumber \\
        &+ d_T \log \left(1 + \frac{T}{\gamma \lambda} \right) \Bigg\},
    \end{align}
    where $\widetilde{H} \defeq N/\min{\{|\mu_i - \tau|^2 : |\mu_i - \tau| \geq \varepsilon\}}$.
    \label{prop:nonadaptive}
\end{proposition}

Thus with a non-adaptive algorithm, the complexity depends only on the most difficult arm (the arm with $\mu_i$ closest to the threshold). As we will see next, with our adaptive approach, the complexity and therefore the algorithmic performance can be significantly improved when there are arms further away from the threshold.

\section{\algoname{}}
\label{sec:grapl}

In this section, we present our algorithm for thresholding graph bandits. Our algorithm is inspired in part by the Anytime Parameter-free Thresholding (APT) algorithm of \citet{locatelli2016optimal}, and also by the work of \citet{valko2014spectral}, who applied Laplacian regularization to the bandit estimator through the eigenvectors of $\L_\lambda$. Unlike \citet{valko2014spectral}, however, we use the Laplacian directly, and we include the tunable regularization parameter $\gamma$. We dub our algorithm the \textbf{Gr}aph-based \textbf{A}nytime \textbf{P}arameter-\textbf{L}ight thresholding algorithm (\textbf{\algoname{}}); see Algorithm~\ref{alg:grapl}. 

\begin{algorithm}
\caption{\algoname{}}
\label{alg:grapl}
\begin{algorithmic}[1]
\State{\textbf{Input:} $\tau, \varepsilon, \L, \gamma, \alpha, \lambda, T$}
\State{$\V_0 \gets \L + \lambda \vec{I}$}
\State{$\widehat{\vec{\mu}}_0 \gets \tau \vec{1}$}
\State{$\widehat{\vec{\Delta}}_0 \gets \varepsilon \vec{1}$}
\State{$\vec{n}_0 \gets \vec{0}$}
\For{$t$ in $1, \ldots, T$}
    \State{$z_i^t \gets \widehat{\Delta}_i^{t-1} \sqrt{n_i^{t-1} + \alpha} \; \forall i$}
    \State{$\pi_t \gets \argmin_i z_i^t$}
    \State{Observe $x_t \sim \nu_{\pi_t}$}
    \State{$\V_t \gets \V_{t-1} + \gamma^{-1} \vec{e}_{\pi_t} \vec{e}_{\pi_t}^\top $}
    \State{$\x_t \gets \x_{t-1} +  \gamma^{-1} x_t \vec{e}_{\pi_t}$}
    \State{$\widehat{\vec{\mu}}_t \gets \V_t^{-1} \x_t$}
    \State{$\widehat{\Delta}_i^t \gets |\widehat{\mu}_i^t - \tau| + \varepsilon \; \forall i$}
    \State{$\vec{n}_t \gets \vec{n}_{t-1} + \vec{e}_{\pi_t}$}
\EndFor
\State{\textbf{Output:} $\widehat{\S} = \left\{ i : \widehat{\mu}_i^T \geq \tau \right\}$}
\end{algorithmic}
\end{algorithm}

At each iteration, \algoname{} performs the same estimation routine as Algorithm~\ref{alg:nonadaptive}. Where it differs is in the strategy for choosing the next arm to sample.
To select the arm at iteration $t+1$, we estimate our distances from the threshold via
\begin{align}
    \widehat{\Delta}_i^t &= |\widehat{\mu}_i^t - \tau| + \varepsilon.
\end{align}
We then use these to compute confidence proxies
\begin{align}
    z_i^{t+1} = \widehat{\Delta}_i^t \sqrt{n_i^t + \alpha},
\end{align}
where $n_i^t$ is the number of times arm $i$ has been selected up to time $t$, and $\alpha > 0$ is some small quantity that keeps $z_i^t$ from being equal to zero before arm $i$ is sampled. Finally, the algorithm selects the next arm~as
\begin{align}
    \pi_t = \argmin_i{z_i^t},
\end{align}
and the next sample is drawn as $x_t \sim \nu_{\pi_t}$. The algorithm then repeats the process in the subsequent iterations until stopped at time $T$.

While \algoname{} has three parameters---namely, $\alpha$, $\lambda$, and $\gamma$---and is therefore not truly parameter-free like APT, the only parameter that needs to be tuned to the specific problem instance is $\gamma$. A value such as $10^{-3}$ for $\lambda$ is sufficient to stabilize the linear system solving in \eqref{eq:muhat} for many problems. If we wish for the algorithm to sample all arms at least once before sampling an arm twice, we can let $\alpha$ be some very small value, such as $10^{-8}$; otherwise, we can let $\alpha$ be a larger value such as $1$. The parameter $\gamma$ is the only parameter that we might wish to choose appropriately based on the graph and the properties of $\vec{\mu}$---see Section~\ref{sec:gamma} for a deeper discussion. However, we note that our main result in Theorem \ref{thm:graphapt} below is valid for \emph{any} values of $\alpha$, $\lambda$, and~$\gamma$.

In terms of implementation, we note that while \eqref{eq:muhat} involves solving a linear system which can be expensive in general, if the graph is sparse, then there exist techniques to solve this system efficiently (in time nearly linear in the number of edges in the graph). Even if the graph is not sparse, it can be ``sparsified'' so that the system can be approximately solved efficiently. We refer the reader to \citet{vishnoi2013lx} for more details. 
We believe this approach (solving the system with $\vec{V}_t$ directly) significantly reduces the complexity of implementing a graph-based bandit algorithm compared to the approach of~\citet{valko2014spectral}, which requires a computation of the eigenspace of $\L_\lambda$. While computing a restricted eigenspace can also be done efficiently using similar techniques, $\algoname{}$ can be implemented in only a few lines of code using a standard solver such as the conjugate gradient method, readily available in common scientific computing packages in most programming languages. We have found such an implementation\footnote{See \url{https://github.com/dlej/grapl}.} fast enough for our purposes when the solver is initialized with the solution from the previous iteration. Though we do not include very large graphs in our experiments in this paper, we have successfully applied \algoname{} to sparse graphs with over 100,000 vertices with no major difficulty. 

\subsection{Error Upper Bounds}

We present a bound on the error that quantifies the extent to which \algoname{} is able to leverage both the graph structure itself and the smoothness of $\vec{\mu}$ on the graph. 

\begin{theorem}
    If Algorithm \ref{alg:grapl} is run on a graph $\mathcal{G}(\mathcal{V}, \mathcal{E}, \vec{W})$ with Laplacian $\L$ and effective dimension $d_T$, and $\|\vec{\mu}\|_{\L_\lambda} \leq \frac{1}{3M+1} \sqrt{\frac{T}{\gamma H}}$, then
    \begin{align}
        \label{eq:grapl-ub}
        \EE{\mathcal{L}_T} \leq \exp \Bigg\{ &- \frac{\gamma^2}{2R^2} \left( \frac{1}{3M+1}\sqrt{\frac{T}{\gamma H}} - \|\vec{\mu}\|_{\L_{\lambda}} \right)^2 \nonumber \\
        &+ d_T \log \left(1 + \frac{T}{\gamma \lambda} \right) \Bigg\},
    \end{align}
    where $M \defeq \max{\left\{\sqrt{\alpha/\gamma\lambda}, \sqrt{1 + \alpha} \right\}}.$
    \label{thm:graphapt}
\end{theorem}

\begin{remark}
    While one must exercise caution when comparing upper bounds, we note that the primary difference between the performance bounds of Algorithm~\ref{alg:nonadaptive} and \algoname{} is in the complexity quantities. The relationship between these two is given by
    \begin{align}
        \widetilde{H} 
        \geq \sum_{i=1}^N \left(\max{\left\{|\mu_i - \tau|, \varepsilon\right\}}\right)^{-2}
        \geq H.
    \end{align}
    That is, in the worst case, where all values $\mu_i$ are close to the threshold $\tau$, we expect both Algorithm~\ref{alg:nonadaptive} and \algoname{} to perform similarly, but when there are only a few values $\mu_i$ near $\tau$, we expect \algoname{} to have a significant advantage.
\end{remark}

\begin{remark}
    We can decompose $T$ as $T = T_0 + T_1$, where
    \begin{align}
        \label{eq:min-T}
        T_0 = \gamma H (3M + 1)^2 \|\vec{\mu}\|_{\L_\lambda}^2
    \end{align}
    is the iteration at which the condition for Theorem~\ref{thm:graphapt} is met, and $T_1$ is the number of iterations after $T_0$. Then for $T_1 \geq 8 T_0$, the right-hand side of \eqref{eq:grapl-ub} can be upper bounded by
    \begin{align}
        \exp \left\{ -\frac{\gamma T_1}{4(3M + 1)^2 R^2 H} 
        + d_{T} \log \left(1 + \frac{T}{\gamma \lambda} \right)
        \right\}.
        \label{eq:ub-simple}
    \end{align}
    While this quantity is controllable by the parameter $\gamma$, this control is limited by the the dependence of $T_0$ on $\gamma$. However, with smaller values of $\|\vec{\mu}\|_{\L_\lambda}$---that is, a \emph{smoother} graph signal---we may realize the faster convergence rates associated with larger values of $\gamma$.
\end{remark}

\begin{remark}
    If we consider the two summands in the exponent of \eqref{eq:ub-simple}, one of the form $-\Theta(T)$ and the other of the form $\Theta(d_T \log T)$, then we can define the \emph{critical iteration} $T_\crit$ as the iteration at which point the first of these terms begins to dominate and the bound begins to rapidly decay with $T$. Specifically, $T_\crit$ is the iteration at which these two terms are equal in magnitude. If we allow the notation $\Thetatilde(\cdot)$ to absorb logarithmic factors, we have that $T_\crit = \Thetatilde(d_T)$. This is already a significant improvement over the standard thresholding bandit problem, where every arm must be drawn at least once, so $T_\crit = \Thetatilde(N)$. 
\end{remark}

\begin{remark}
    \label{remark:offset}
    The quantity $\|\vec{\mu}\|_{\L_\lambda}$ can be considered with respect to any reference offset used to estimate $\widehat{\vec{\mu}}$. For example, if we replaced \eqref{eq:muhat} and \eqref{eq:x} with
    \begin{align}
        \widehat{\vec{\mu}}_t &= \V_t^{-1} \x_t + \tau \vec{1} \\
        \x_t &= \frac{1}{\gamma}\sum_{s=1}^t (x_s - \tau) \vec{e}_{\pi_s},
    \end{align}
    then the $\|\vec{\mu}\|_{\L_\lambda}$ quantities in the above bound would be replaced by $\|\vec{\mu} - \tau \vec{1}\|_{\L_\lambda}$.
\end{remark}

\subsection{Optimality}

\subsubsection{Oracle Sampling Strategy}

Consider an oracle algorithm that uses the same estimation strategy as Algorithm~\ref{alg:nonadaptive} and \algoname{} but has access to the values of $|\mu_i - \tau|$ and need only identify the sign of $\mu_i - \tau$. Instead of a non-adaptive sampling strategy, let this algorithm sample according to its knowledge of $|\mu_i - \tau|$. For such an algorithm, if we relax the notion of sampling to allow the algorithm to make non-integer sample allocations according to an allocation rule $\vec{\beta}$ (obeying $\beta_i \geq 0$ and $\sum_i \beta_i = 1$) such that $n_i^t = \beta_i t$, we obtain the following result.

\begin{proposition}
    For the oracle algorithm with sampling allocation $\vec{\beta}$, if $\|\vec{\mu}\|_{\L_\lambda} \leq \sqrt{\frac{T}{\gamma H_*}}$, then 
    \begin{align}
        \label{eq:oracle-ub}
        \inf_{\vec{\beta}} \EE{\mathcal{L}_T} \leq \exp \Bigg\{ &- \frac{\gamma^2}{2R^2} \left( \sqrt{\frac{T}{\gamma H_*}} - \|\vec{\mu}\|_{\L_{\lambda}} \right)^2 \nonumber \\
        &+ d_T \log \left(1 + \frac{T}{\gamma \lambda} \right) \Bigg\},
    \end{align}
    where
    $H_* \defeq \sum_{j : |\mu_j - \tau| \geq \varepsilon} |\mu_j - \tau|^{-2}$.
    \label{prop:oracle}
\end{proposition}

We note the similarity between $H$ and $H_*$. Using the fact that $|\mu_i - \tau| + \varepsilon \leq 2 |\mu_i - \tau|$ for $|\mu_i - \tau| \geq \varepsilon$, we can relate the two by
\begin{align}
    4H \geq H_* \geq H - \varepsilon^{-2} N_\mathrm{small},
\end{align}
where $N_\mathrm{small} = \left| \left\{ i : |\mu_i - \tau| < \varepsilon \right\} \right|$.
So, except in cases where there are many values $\mu_i$ that are near the threshold, the performance upper bound of \algoname{} matches that of the oracle algorithm. However, in cases where there are many values $\mu_i$ that are within $\varepsilon$ of the threshold, the oracle algorithm can have significantly lower complexity.

\subsubsection{Lower Bound for Disconnected Cliques}

Consider the following family of graphs of size $N$ consisting of $D$ disconnected $K$-cliques and associated graph signals $\vec{\mu}$ such that for each arm $i$ belonging to clique $j$, $\mu_i=\mu_j$. For this family of graphs and signals, the thresholding graph bandit problem reduces  to the thresholding bandit problem on $D$ independent arms with complexity $H' \defeq \sum_{j=1}^D (|\mu_j - \tau| + \varepsilon)^{-2} = H/K$. This gives us the following lower bound from \citet{locatelli2016optimal}:
\begin{align}
    \EE{\mathcal{L}_T} \geq \exp \left\{ - \frac{3 K T}{R^2 H} - 4 \log(12(\log(T) + 1)N) \right\}.
\end{align}
For the lower bound, then, $T_\crit = \Thetatilde(R^2 H / K)$.

For this family of graphs, the graph Laplacian consists of a matrix with $D$
blocks of the form $K\vec{I}_K - \vec{J}_K$, where $\vec{J}_K$ is the $K \times K$ matrix of all ones. Therefore, the eigenvalues of $\L_\lambda$ are $\lambda$ with multiplicity $D$ and $K + \lambda$ with multiplicity $N - D$. Thus, the effective dimension is the larger of
\begin{align*}
    \min \left\{D, \floor*{1 + \frac{T}{\gamma \lambda \log(1 + T/\gamma \lambda)}} \right\}
\end{align*}
and 
\begin{align*}
    \min \left\{ N, \floor*{1 + \frac{T}{\gamma (K + \lambda) \log(1 + T/\gamma \lambda)}} \right\}.
\end{align*}
For any desired time horizon (e.g., $T \leq 10,000$), for sufficiently small $\lambda$, this will result in $d_T \leq D$. We also note that for this class of signals, $\|\vec{\mu}\|_{\L_\lambda}^2 = \lambda \|\vec{\mu}\|_2^2$, so for sufficiently small $\lambda$, the bound in Theorem~\ref{thm:graphapt} holds for all $T$.

Considering the form of our upper bound in \eqref{eq:ub-simple}, we have for this problem class that $T_\crit = \Thetatilde(D R^2 H / \gamma) = \Thetatilde(N R^2 H / \gamma K)$. So, considering a fixed $N$ and $\gamma$, we can say that \algoname{} has optimal $T_\crit$ (up to logarithmic factors) with respect to $R$, $H$, and $K$ (equivalently, $D$) for this family of graphs and signals. With $\gamma=N$, this rate would also be optimal with respect to $N$ if it were not for the condition in \eqref{eq:min-T}. 

\subsubsection{Linear Bandits}
As pointed out by \citet{valko2014spectral}, if $\vec{\mu}$ lies in the span of $D$ eigenvectors of $\L$, then the graph bandit problem reduces to the problem of thresholding \emph{linear bandits} \citep{Auer:2003:UCB:944919.944941}. Results from the \emph{best arm identification} problem in linear bandits \citep{NIPS2014_5460,pmlr-v80-tao18a}, another example of pure exploration bandits, suggest that the optimal sample complexity is linear in the underlying dimension $D$. In the above example with graphs consisting of $D$ cliques, signal $\vec{\mu}$ lies in the span of the $D$ eigenvectors corresponding to the smallest eigenvalues of $\L$, and so our result that $T_\crit = \Thetatilde(D)$ in this setting is consistent with results from linear bandits.

\subsection{Choice of Regularization Parameter}
\label{sec:gamma}

\algoname{} has a free parameter $\gamma$ which can be tuned to optimize $T_\crit$, which we discuss in this section.
$T_\crit$ will be on the order of the larger of $T_0$ and $T_1$, so to optimize $T_\crit$, we must fix $T_0$ and $T_1$ to be of the same order. Here, we simply set $T_1 = 8 T_0$. 
Then our optimal choice of $\gamma$ is that which satisfies \eqref{eq:min-T} and
\begin{align*}
    \frac{\gamma T_1}{4 (3M + 1)^2 R^2 H} = d_{T_0 + T_1} \log \left(1 + \frac{T_0 + T_1}{\gamma \lambda} \right).
\end{align*}
After some algebra, we obtain
\begin{align}
    \gamma^* = \frac{2R}{\|\vec{\mu}\|_{\L_\lambda}} \sqrt{d' \log \left(1 + \frac{9H(3M + 1)^2 \|\vec{\mu}\|_{\L_\lambda}^2}{\lambda} \right)},
    \label{eq:gamma-star}
\end{align}
where $d'$, the effective dimension at time $T_0 + T_1$ for this choice of $\gamma$, is the largest $d$ such that
\begin{align*}
    (d - 1)\lambda_d \leq \frac{9H(3M + 1)^2 \|\vec{\mu}\|_{\L_\lambda}^2}{\log \left(1 + \frac{9H(3M + 1)^2 \|\vec{\mu}\|_{\L_\lambda}^2}{\lambda} \right)}.
\end{align*}
As we would expect, the smoother the graph signal is (smaller $\|\vec{\mu}\|_{\L_\lambda}$) and the larger the amount of noise, the larger $\gamma$ (the more smoothing) we will require. All together, this gives us $T_\crit = \Thetatilde(\sqrt{d'}R \|\vec{\mu}\|_{\L_\lambda} H)$. In the worst case, when the graph structure is unhelpful (i.e., when $d' = N$) and the signal is not smooth on the graph (i.e., $\|\vec{\mu}\|_{\L_\lambda} = \Theta(\sqrt{N})$), this gives $T_\crit$ a linear dependence on $N$, as we would expect. On the other hand, in the setting of $D$ cliques, where for sufficiently small $\lambda$ we can consider $\|\vec{\mu}\|_{\L_\lambda} = \Theta(\sqrt{D})$, we again obtain $T_\crit = \Thetatilde(D)$.

\section{EXPERIMENTS}
\label{sec:experiments}
In experiments on both artificial and real data we demonstrate the advantage of \algoname{} over the APT algorithm of~\citet{locatelli2016optimal}, which does not utilize the graph information, and over Algorithm~\ref{alg:nonadaptive}, which uses non-adaptive random arm sampling. We demonstrate that exploiting the graph structure can significantly reduce the number of samples necessary to obtain a good estimate of the superlevel set, and that the adaptive arm selection rule of \algoname{} further reduces the number of samples necessary over non-adaptive sampling with the same graph-regularized estimator.

\subsection{Stochastic Block Model}

\begin{figure}[t]
    \centering
    \includegraphics[width=0.9\linewidth]{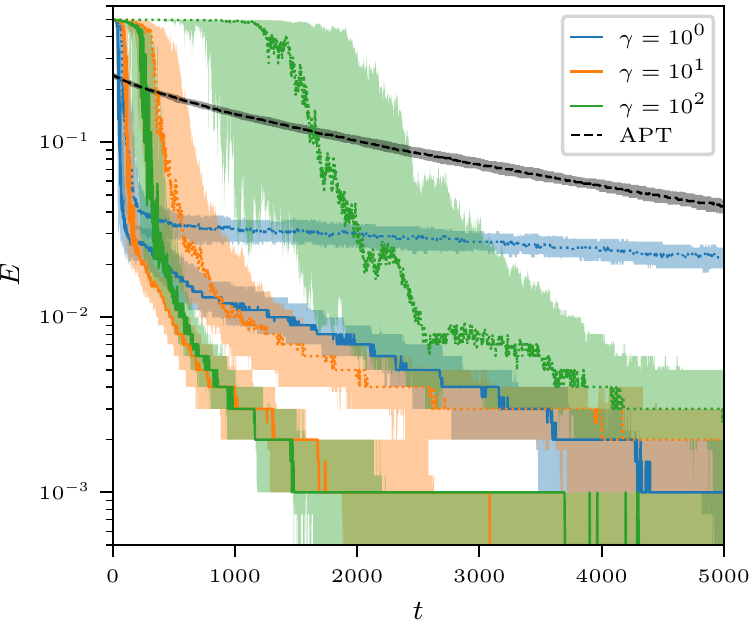}
    \caption{Misclassification error $E$ vs. iteration $t$ on the stochastic block model problem for \algoname{} (solid), APT (dashed), and Algorithm~\ref{alg:nonadaptive} (dotted). Lines indicate the median error, and shaded areas around the lines indicate the interquartile range. Solid and dotted lines of the same color use the same value of $\gamma$ for \algoname{} and Algorithm~\ref{alg:nonadaptive}, respectively.}
    \label{fig:sbm}
\end{figure}

In our first experiment, we let $N = 1000$ and sample an unweighted, undirected graph from a stochastic block model with two communities of size $N/2$, with within-community edge probability $\log(N/2) / (N/2)$ and between-community edge probability $\log(N/2) / (N/2)^{3/2}$. We let
\begin{align*}
    \mu_i = \begin{cases}
    1 & i \leq N/2 \\
    -1 & \text{otherwise},
    \end{cases}
\end{align*}
and we make the distribution of each arm Gaussian with $\sigma = 2$. For \algoname{}, we let $\lambda = 10^{-3}$ and $\alpha = 1$. With $\tau = 0$ and $\varepsilon = 0.01$, we run the algorithms for $T=5000$ iterations and compute the misclassification error $E$ at each iteration $t$, defined as
\begin{align}
    E = \frac{\left| (\S_{\tau + \varepsilon} \cap \widehat{\S}^c) \cup (\S_{\tau - \varepsilon}^c \cap \widehat{\S}) \right|}{\left|  \S_{\tau + \varepsilon} \cup \S_{\tau - \varepsilon}^c \right|}.
\end{align}
Figure \ref{fig:sbm} shows the median misclassification error for each algorithm and choice of $\gamma$ over $100$ trials along with the interquartile range. We note that APT is initialized with an additional $2N = 2000$ samples before its first iteration, so for APT the actual number of samples collected is higher than the iteration counter. Both \algoname{} and Algorithm~\ref{alg:nonadaptive} (for sufficiently large $\gamma$) are able to exploit the graph structure and converge to the correct superlevel set much more quickly than APT. However, consistently across values of $\gamma$, \algoname{} converges in turn much more quickly than its non-adaptive counterpart. In particular, \algoname{} makes significant gains in early iterations and appears to be more robust to the choice of $\gamma$. We also computed $\gamma^*$ according to \eqref{eq:gamma-star} for this problem and found the average $\gamma^*$ to be 28.72 with a standard deviation of 1.15 over 100 trials, which agrees with the good performance of \algoname{} with $\gamma = 10$ and $\gamma = 100$.

\subsection{Small-World Graph}

In our next experiment, we again let $N = 1000$ and sample small-world graphs according to the model of \citet{newman1999renormalization} with new-edge probability $0.01$ and ring initialized with $4$ neighbors. To generate our smooth signal, we first generate an i.i.d. Gaussian vector $\y \in \R^N$ and compute
\begin{align*}
    \vec{\mu}_0 = (\L + \vec{I} / N^2)^{-1} \y,
\end{align*}
which we then normalize to have zero median and standard deviation 0.2. The multiplication by $(\L + \vec{I} / N^2)^{-1}$ serves essentially to project $\vec{y}$ onto the eigenspace of $\L$ corresponding to its smallest eigenvalues, which vary smoothly along the graph. Following this, we obtain $\vec{\mu}$ by adding 0.5 to the signal and clipping the values to be between 0 and 1. The distribution of each arm is Bernoulli with probability $\mu_i$. With $\tau = 0.5$ and $\varepsilon = 0.01$, the problem is quite difficult.

\begin{figure}[t]
    \centering
    \includegraphics[width=0.9\linewidth]{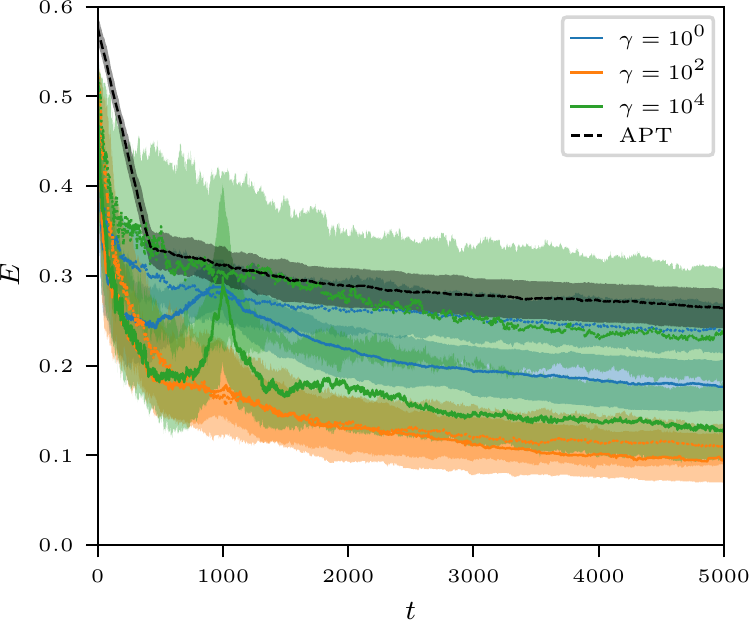}
    \caption{Misclassification error $E$ vs iteration $t$ on the small-world graph problem for \algoname{} (solid), APT (dashed), and Algorithm~\ref{alg:nonadaptive} (dotted). Lines indicate the median error, and shaded areas around the lines indicate the interquartile range. Solid and dotted lines of the same color use the same value of $\gamma$ for \algoname{} and Algorithm~\ref{alg:nonadaptive}, respectively.}
    \label{fig:newman-watts}
\end{figure}

Figure~\ref{fig:newman-watts} shows the misclassification error for this problem when the algorithms are run over 100 trials for $T = 5000$ iterations. As before, we show the median error and interquartile range. For \algoname{}, we let $\lambda = 10^{-3}$ and $\alpha = 10^{-8}$, and we estimate $\widehat{\vec{\mu}}$ with respect to the offset $\tau$ as described in Remark~\ref{remark:offset}. On this much more difficult problem, we have selected a wider range of values for $\gamma$. Here we again see that although with the best choice of $\gamma$ the advantage of \algoname{} is only slight over Algorithm~\ref{alg:nonadaptive}, \algoname{} is much more robust to the choice of $\gamma$, and for poorly chosen $\gamma$ the non-adaptive algorithm provides almost no advantage over APT. We found the average $\gamma^*$ to be 227.9 with a standard deviation of 50.9 over 100 trials for this problem, which agrees with our finding the best performance at $\gamma = 100$. Lastly, we note that an artifact of the choice of very small $\alpha$ is that there is a spike in error around $t = N$ which corresponds to \algoname{} prioritizing sampling each arm at least once over the adaptive strategy.

\subsection{Political Blogs}

In our experiment on real-world data, we use the political blogs graph from \citet{adamic2005political}. The vertices in the graph correspond to political blogs commenting on US politics around the time of the 2004 U.S. presidential campaign, and edges denote links from one blog to another. The signal $\vec{\mu}$ associated with this graph is 
\begin{align*}
    \mu_i = \begin{cases}
    1 & \text{blog $i$ is conservative-leaning} \\
    0 & \text{blog $i$ is liberal-leaning}.
    \end{cases}
\end{align*}
We make the edges undirected and set the edge weight equal to the total number of links from one blog to the other, and then take the largest connected component, which contained 1222 blogs. The problem we simulate then is that we would like to identify which of these blogs are conservative and liberal without actually having to visit and read each blog (expensive sampling), and we have access to this additional graph information (and cheap computation compared to the time it would take to visit a blog). We make the distribution of each arm non-random and let the algorithms take at most $N$ samples. Since APT requires $2N$ samples for initialization, we do not compare against APT.

\begin{figure}[t]
    \centering
    \includegraphics[width=0.9\linewidth]{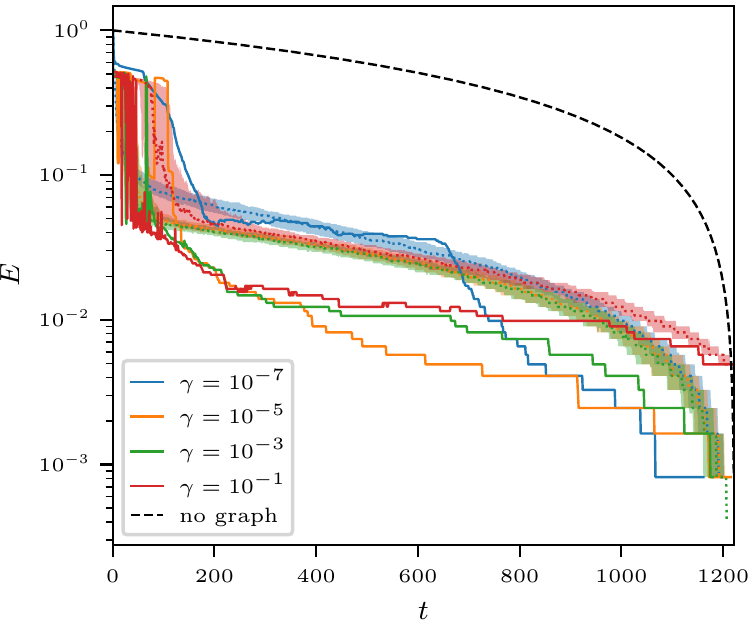}
    \caption{Misclassification error $E$ vs iteration $t$ on the political blogs problem for \algoname{} (solid), Algorithm~\ref{alg:nonadaptive} (dotted), and using no graph (dashed). For Algorithm~\ref{alg:nonadaptive}, lines indicate median error, and shaded areas around the lines indicate the interquartile range. Solid and dotted lines of the same color use the same value of $\gamma$. }
    \label{fig:polblogs}
\end{figure}

Figure \ref{fig:polblogs} shows the misclassification error for $\tau = 0.5$ and $\varepsilon = 0.01$, with median error and interquantile range over 100 trials for Algorithm~\ref{alg:nonadaptive}. We run \algoname{} with $\lambda = 10^{-3}$ and $\alpha = 10^{-8}$, using offset $\tau$, and vary $\gamma$, but we do not run repeated trials since the observations are non-random. The results are similar to before, in that using the graph structure provides much better results than not using the graph, and in that we see \algoname{} consistently outperforming Algorithm~\ref{alg:nonadaptive}.  For instance, with $\gamma = 10^{-5}$, \algoname{} is able to reach 1\% error at $t \approx 400$, while its random counterpart over the majority of trials does not do the same until $t > 1000$. We would expect the optimal $\gamma$ to be the smallest $\gamma$ possible based on \eqref{eq:gamma-star}, since there is no noise in the problem. However, for $\gamma=10^{-7}$, floating point rounding begins to become an issue---effectively, there is a small nonzero amount of noise due to rounding---and the performance of \algoname{} is worse than with the larger values of $\gamma$.

\section{CONCLUDING REMARKS}

In this paper we have introduced a new paradigm of online sequential decision making that we call Thresholding Graph Bandits, where the main objective is the identification of the superlevel set of arms whose means are above a given threshold in a multi-armed bandit setting. Importantly, in our framework, we have supposed that we have access to a graph that encodes the similarity between the arms. We have developed \algoname{}, a novel algorithm for this thresholding graph bandits problem, along with theoretical results that show the relationship between the misclassification rate of \algoname{}, the number of arm pulls, the graph structure, and the smoothness of the reward function with respect to the given graph. We have also demonstrated that \algoname{} is optimal in terms of the number of arm pulls, the statistical hardness, and the dimensionality of the problem. Finally, we have confirmed our theoretical results via experiments on synthetic and real data, highlighting the significant gains to be had in leveraging the graph information with an adaptive algorithm. 

\subsubsection*{Acknowledgements}
This work was supported by 
NSF grants CCF-1911094, IIS-1838177, and IIS-1730574; 
ONR grants N00014-18-12571 and N00014-17-1-2551;
AFOSR grant FA9550-18-1-0478; 
DARPA grant G001534-7500; and a 
Vannevar Bush Faculty Fellowship, ONR grant N00014-18-1-2047.

\appendix

\onecolumn

\section{USEFUL LEMMAS}

We introduce the additional notation of
\begin{gather}
    \vec{\xi}_t = \sum_{s=1}^t \vec{e}_{\pi_s} (x_s - \mu_{\pi_s}), \\
    \sigma_i^t = \sqrt{(\V_t^{-1})_{ii}}, \\
    \vec{N}_t = \mathrm{diag}(\vec{n}_t),
\end{gather}
to be used in the proofs of our results. The following lemmas are proved in Section~\ref{sec:lemma-proofs}.

\begin{lemma}
    \label{lemma:mean-diff}
    With probability at least $1-\delta$, for any $i \in [N]$ and $t \geq 1$, 
    \begin{equation}
        |\widehat{\mu}_i^t - \mu_i| \leq \sigma_i^t \left( \frac{R}{\gamma} \sqrt{\log \left(\frac{|\V_t|}{\delta^2 |\L_{\lambda}|} \right)} + \|\vec{\mu}\|_{\L_{\lambda}} \right).
    \end{equation}
\end{lemma}

\begin{lemma}
    \label{lemma:sigma-bound}
    For all $i \in [N]$ and $t \geq 0$,
    \begin{equation}
        \sigma_i^t \leq \sqrt{\frac{(\sigma_i^0)^2}{1 + (\sigma_i^0)^2 n_i^t / \gamma}}.
    \end{equation}
\end{lemma}

\begin{lemma}
    \label{lemma:log-det}
    Let $d_T$ be the effective dimension. Then
    \begin{equation}
        \log \frac{|\V_T|}{|\L_\lambda|} \leq 2d_T \log \left( 1 + \frac{T}{\gamma \lambda} \right).
    \end{equation}
\end{lemma}

\section{PROOF OF PROPOSITION~\ref{prop:nonadaptive}}

For Algorithm~\ref{alg:nonadaptive} to succeed, it must be that $\widehat\mu_i \geq \tau$ for each $i$ such that $\mu_i \geq \tau + \varepsilon$ and $\widehat\mu_i < \tau$ for each $i$ such that $\mu_i < \tau - \varepsilon$ (we can make this inequality strict or non-strict without changing probabilistic statements since $\widehat{\vec{\mu}}$ is a continuous random variable). For a given $i$, this is satisfied if $|\widehat{\mu}_i - \mu_i| \leq |\mu_i - \tau|$. We show this for the case that $\mu_i \geq \tau + \varepsilon$. If $\widehat{\mu}_i \geq \mu_i$ in this case, then the necessary condition is satisfied. If $\widehat{\mu}_i < \mu_i$, then
\begin{align}
    \mu_i - \tau = |\mu_i - \tau| &\geq |\widehat{\mu}_i - \mu_i|
    = \mu_i - \widehat{\mu}_i \\
    \implies \tau &\leq \widehat{\mu}_i.
\end{align}
The case where $\mu_i \leq \tau - \varepsilon$ is analogous. Thus, a sufficient condition for the success of Algorithm~\ref{alg:nonadaptive} is that $|\widehat{\mu}_i - \mu_i| \leq |\mu_i - \tau|$ for all $i$ such that $|\mu_i - \tau| \geq \varepsilon$. If we use Lemmas~\ref{lemma:mean-diff}, \ref{lemma:sigma-bound}, and \ref{lemma:log-det}, we know that with probability at least $1 - \delta$,
\begin{align}
    |\widehat{\mu}_i^t - \mu_i| &\leq \sigma_i^t \left( \frac{R}{\gamma} \sqrt{\log \left(\frac{|\V_t|}{\delta^2 |\L_{\lambda}|} \right)} + \|\vec{\mu}\|_{\L_{\lambda}} \right) \\
    &\leq \sqrt{\frac{(\sigma_i^0)^2}{1 + (\sigma_i^0)^2 n_i^t / \gamma}} \left( \frac{R}{\gamma} \sqrt{2d_T \log \left( 1 + \frac{T}{\gamma \lambda} \right) - 2 \log \delta} + \|\vec{\mu}\|_{\L_{\lambda}}
    \right)
    \\
    &\leq \sqrt{\frac{\gamma}{n_i^t}} \left( \frac{R}{\gamma} \sqrt{2d_T \log \left( 1 + \frac{T}{\gamma \lambda} \right) - 2 \log \delta} + \|\vec{\mu}\|_{\L_{\lambda}}
    \right).
\end{align}
Thus Algorithm~\ref{alg:nonadaptive} succeeds with probability at least $1 - \delta$ if, for all $i$ such that $|\mu_i - \tau| \geq \varepsilon$,
\begin{align}
    \label{eq:all-i-bound}
    \sqrt{\frac{\gamma}{n_i^t}} \left( \frac{R}{\gamma} \sqrt{2d_T \log \left( 1 + \frac{T}{\gamma \lambda} \right) - 2 \log \delta} + \|\vec{\mu}\|_{\L_{\lambda}}
    \right) \leq |\mu_i - \tau|.
\end{align}
Because Algorithm~\ref{alg:nonadaptive} has an equal sampling allocation for each arm, for $T=kN$ we have that $n_i^t = k = T/N$. Then since for each $i$ the left-hand side of \eqref{eq:all-i-bound} is the same, we can write the complete sufficient condition as
\begin{align}
    \sqrt{\frac{\gamma N}{T}} \left( \frac{R}{\gamma} \sqrt{2d_T \log \left( 1 + \frac{T}{\gamma \lambda} \right) - 2 \log \delta} + \|\vec{\mu}\|_{\L_{\lambda}}
    \right) \leq \min \left\{|\mu_i - \tau| : |\mu_i - \tau| \geq \varepsilon \right\}.
\end{align}
The smallest $\delta$ for which this inequality holds is
\begin{align}
    \delta = \exp \Bigg\{ &- \frac{\gamma^2}{2R^2} \left( \sqrt{\frac{T}{\gamma \widetilde{H}}} - \|\vec{\mu}\|_{\L_{\lambda}} \right)^2 
        + d_T \log \left(1 + \frac{T}{\gamma \lambda} \right) \Bigg\},
\end{align}
provided $\|\vec{\mu}\|_{\L_\lambda} \leq \sqrt{\frac{T}{\gamma \widetilde{H}}}$, where $\widetilde{H} \defeq N/\min{\{|\mu_i - \tau|^2 : |\mu_i - \tau| \geq \varepsilon\}}$.

\section{PROOF OF THEOREM~\ref{thm:graphapt}}

The proof follows the same general strategy as that of Theorem 2 of \citet{locatelli2016optimal}. 
\subsection{A Favorable Event}

Let 
\begin{align}
\delta = \exp \bigg\{ &- \frac{\gamma^2}{2R^2} \left( \frac{1}{3M+1}\sqrt{\frac{T}{\gamma H}} - \|\vec{\mu}\|_{\L_{\lambda}} \right)^2 
+ d_T \log \left(1 + \frac{T}{\gamma \lambda} \right) \bigg\},
\end{align}
and consider for the rest of the proof an event of probability at least $1 - \delta$ that gives us the result of Lemma \ref{lemma:mean-diff}. On this event then, for all $i \in [N]$,
\begin{align}
    |\widehat{\mu}_i^t - \mu_i| 
    &\leq \sigma_i^t \left( \frac{R}{\gamma} \sqrt{\log \left(\frac{|\V_t|}{\delta^2 |\L_{\lambda}|} \right)}  + \|\vec{\mu}\|_{\L_{\lambda}} \right) 
    \nonumber \\
    &\leq \sigma_i^t \left(\frac{R}{\gamma}\sqrt{2d_T\log (1 + T/\gamma\lambda) - 2\log \delta} + \|\vec{\mu}\|_{\L_{\lambda}} \right) \nonumber \\
    &\leq \frac{\sigma_i^t}{3M + 1} \sqrt{\frac{T}{\gamma H}},
\end{align}
where the second inequality comes from Lemma \ref{lemma:log-det} and the third inequality comes from plugging in $\delta$ using the fact that $\|\vec{\mu}\|_{\L_{\lambda}} \leq \frac{1}{3M+1} \sqrt{\frac{T}{\gamma H}}$.

\subsection{A Helpful Arm}

At time $T$, there must exist an arm $k$ such that $n_k^T \geq \frac{T} {H \Delta_k^2}$. If this were not true, then
\begin{align}
    T = \sum_{i=1}^N n_i^T < \sum_{i=1}^N \frac{T} {H \Delta_i^2} = T,
\end{align}
which is a contradiction. Let $t \leq T$ be the last time this arm was pulled, and consider this time for the rest of the proof. Note that $n_k^t = n_k^T \geq  \frac{T} {H \Delta_k^2}$.

\subsection{Bounding the Other Arms using the Helpful Arm}

When $n_i^t \geq 1$, using Lemma \ref{lemma:sigma-bound},
\begin{align}
    \label{eq:sigma-n-geq1-bound}
    \sigma_i^t \sqrt{n_i^t + \alpha} &\leq \sqrt{\frac{(\sigma_i^0)^2 (n_i^t + \alpha)}{1 + (\sigma_i^0)^2 n_i^t / \gamma}} \nonumber \\
    &\leq \sqrt{\frac{\gamma (n_i^t + \alpha)}{n_i^t}} \nonumber \\
    &\leq \sqrt{\gamma (1 + \alpha)}.
\end{align}
So, including the case of $n_i^t = 0$, 
\begin{align}
    \label{eq:sigma-n-bound}
    \sigma_i^t \sqrt{n_i^t + \alpha} &\leq \max{\left\{\sigma_i^0 \sqrt{\alpha}, \sqrt{\gamma (1 + \alpha)} \right\}} \leq \sqrt{\gamma} M,
\end{align}
where the last inequality comes from the fact that ${\sigma_i^0 \leq 1/\sqrt{\lambda}}$.

We know that
\begin{align}
     |\widehat{\mu}_i^t - \mu_i| &\geq  \left| |\widehat{\mu}_i^t - \tau| - |\mu_i - \tau| \right|
     = | \widehat{\Delta}_i^t - \Delta_i |,
\end{align}
so we can find a lower bound:
\begin{align}
    z_k^t &= \widehat{\Delta}_k^t \sqrt{n_k^t + \alpha} \nonumber \\
    &\geq \left( \Delta_k - \frac{\sigma_k^t}{3M + 1} \sqrt{\frac{T}{\gamma H}} \right) \sqrt{n_k^t} \nonumber \\
    &\geq \sqrt{\frac{T}{H}} \frac{3M}{3M + 1},
\end{align}
where the last inequality comes from our bound on $n_k^t$ and from (\ref{eq:sigma-n-geq1-bound}) with $\alpha = 0$. For the upper bound,
\begin{align}
    z_i^t &= \widehat{\Delta}_i^t \sqrt{n_i^t + \alpha} \nonumber \\
    &\leq \left( \Delta_i + \frac{\sigma_i^t}{3M + 1} \sqrt{\frac{T}{\gamma H}} \right) \sqrt{n_i^t + \alpha} \nonumber \\
    &\leq \Delta_i \sqrt{n_i^t + \alpha} + \frac{M}{3M + 1} \sqrt{\frac{T}{H}}.
\end{align}
Since we pulled arm $k$ on round $t$, $z_k^t \leq z_i^t$, so
\begin{gather}
    \sqrt{\frac{T}{H}} \frac{3M}{3M + 1} \leq \Delta_i \sqrt{n_i^t + \alpha} + \frac{M}{3M + 1} \sqrt{\frac{T}{H}}, \\
    \label{eq:delta-bound}
    \implies
    \frac{1}{3M + 1} \sqrt{\frac{T}{H}} \leq \frac{\Delta_i \sqrt{n_i^t + \alpha}}{2M}.
\end{gather}

\subsection{Wrapping Up}
Finally, we have that
\begin{align}
    |\widehat{\mu}_i^T - \mu_i| &\leq \frac{\sigma_i^T}{3M + 1} \sqrt{\frac{T}{\gamma H}}
    \leq \frac{\Delta_i \sigma_i^t \sqrt{n_i^t + \alpha}}{2\sqrt{\gamma}M} 
    \leq \frac{\Delta_i}{2},
\end{align}
where the second inequality comes from the fact that $\sigma_i^t$ is decreasing in $t$ and from (\ref{eq:delta-bound}).
Now for $i$ such that $\mu_i \geq \tau + \varepsilon$, we have
\begin{align}
    \widehat{\mu}_i^T &\geq \mu_i - \frac{\Delta_i}{2} 
    = \mu_i - \frac{\mu_i - \tau + \varepsilon}{2} 
    = \frac{\tau + \mu_i - \varepsilon}{2}
    \geq \tau.
\end{align}
For $i$ such that $\mu_i \leq \tau - \varepsilon$, we have
\begin{align}
    \widehat{\mu}_i^T &\leq \mu_i + \frac{\Delta_i}{2} 
    = \mu_i + \frac{\tau - \mu_i + \varepsilon}{2} 
    = \frac{\tau + \mu_i + \varepsilon}{2}
    \leq \tau.
\end{align}

\section{PROOF OF PROPOSITION~\ref{prop:oracle}}

The proof of this proposition is the same as the proof of proposition~\ref{prop:nonadaptive} until the choice of the sampling allocation $n_i^t = \beta_i t$. Continuing from \eqref{eq:all-i-bound}, we must choose $\vec{\beta}$ such that, for all $i$ such that $|\mu_i - \tau| \geq \varepsilon$,
\begin{align}
    \sqrt{\frac{\gamma}{T}} \left( \frac{R}{\gamma} \sqrt{2d_T \log \left( 1 + \frac{T}{\gamma \lambda} \right) - 2 \log \delta} + \|\vec{\mu}\|_{\L_{\lambda}}
    \right) \leq \sqrt{\beta_i} |\mu_i - \tau|.
\end{align}
To optimize this inequality such that it holds for the smallest possible $\delta$, we must make the right-hand side as large as possible. That is, we must choose $\vec{\beta}$ that maximizes
\begin{align}
\min_{i : |\mu_i - \tau| \geq \varepsilon}
\sqrt{\beta_i} |\mu_i - \tau|.
\end{align}
To maximize this minimum, we must choose $\vec{\beta}$ that makes all of the terms the same. With the constraint that $\sum_i \beta_i = 1$, this means that we must choose
\begin{align}
    \beta_i = 
    \begin{cases}
        \left(H_* |\mu_i - \tau|^2\right)^{-1} & \text{if } |\mu_i - \tau| \geq \varepsilon \\
        0 & \text{otherwise},
    \end{cases}
\end{align}
where
\begin{align}
    H_* = \sum_{j : |\mu_j - \tau| \geq \varepsilon} |\mu_j - \tau|^{-2}.
\end{align}
With this choice of $\vec{\beta}$, the smallest $\delta$ for which the inequality holds is
\begin{align}
    \delta = \exp \Bigg\{ &- \frac{\gamma^2}{2R^2} \left( \sqrt{\frac{T}{\gamma H_*}} - \|\vec{\mu}\|_{\L_{\lambda}} \right)^2 
        + d_T \log \left(1 + \frac{T}{\gamma \lambda} \right) \Bigg\},
\end{align}
provided $\|\vec{\mu}\|_{\L_\lambda} \leq \sqrt{\frac{T}{\gamma H_*}}$.

\section{PROOF OF LEMMAS}
\label{sec:lemma-proofs}

\subsection{Proof of Lemma~\ref{lemma:mean-diff}}

To prove Lemma~\ref{lemma:mean-diff}, we first need the following lemma, which is a direct consequence of Theorem 1 of \citet{abbasi2011improved}:

\begin{lemma}
    \label{lemma:self-normalized}
    For any $\delta > 0$, with probability at least $1 - \delta$, for all $t \geq 0$,
    \begin{equation}
    \|\vec{\xi}_t\|_{V_t^{-1}}^2 \leq R^2 \log \left( \frac{|\V_t|}{\delta^2 |\L_{\lambda}|}\right).
    \end{equation}
\end{lemma}

Using Lemma~\ref{lemma:self-normalized}, the proof of Lemma~\ref{lemma:mean-diff} follows that of Lemma 3 of \citet{valko2014spectral}.
    Let $\vec{N}_t = \mathrm{diag}(\vec{n}_t)$, and note that $\x_t = {(\vec{N}_t\vec{\mu} + \vec{\xi}_t)/\gamma}$.
    Then
    \begin{align}
        |\widehat{\mu}_i^t - \mu_i| &= \left| \langle \vec{e}_i, \V_t^{-1} (\vec{N}_t\vec{\mu} + \vec{\xi}_t)/\gamma - \vec{\mu} \rangle\right| \nonumber \\
        &= \left| \langle \vec{e}_i, \V_t^{-1} \vec{\xi}_t / \gamma - \V_t^{-1} \left( \V_t - \vec{N}_t /\gamma \right) \vec{\mu} \rangle \right| \nonumber \\
        &\leq \left| \langle \vec{e}_i, \vec{\xi}_t / \gamma \rangle_{\V_t^{-1}} \right|
        + \left| \langle \vec{e}_i, \L_{\lambda} \vec{\mu} \rangle_{\V_t^{-1}} \right| \nonumber \\
        &\leq \sigma_i^t \left( \|\vec{\xi}_t / \gamma\|_{\V_t^{-1}} + \|\L_{\lambda} \vec{\mu}\|_{\V_t^{-1}}\right),
    \end{align}
    where the last inequality comes from Cauchy-Schwarz and the fact that $\sigma_i^t = \|\vec{e}_i\|_{\V_t^{-1}}$. The first term is bounded by Lemma \ref{lemma:self-normalized}, and the second term is bounded as follows:
    \begin{align}
        \|\L_{\lambda} \vec{\mu}\|_{\V_t^{-1}}^2
        &= \vec{\mu}^\top \L_{\lambda} \V_t^{-1} \L_{\lambda} \vec{\mu} \nonumber \\
        &= \vec{\mu}^\top \left( \L_{\lambda} - \vec{N}_t^{1/2} \left(\gamma \vec{I} + \vec{N}_t^{1/2} \L_{\lambda} \vec{N}_t^{1/2}\right)^{-1} \vec{N}_t^{1/2} \right)  \vec{\mu} \nonumber \\
        &\leq \vec{\mu}^\top \L_{\lambda} \vec{\mu} = \|\vec{\mu}\|_{\L_{\lambda}}^2,
    \end{align}
    where the second equality comes from the Woodbury matrix identity, and the first inequality is from the subtrahend being positive semidefinite.

\subsection{Proof of Lemma~\ref{lemma:sigma-bound}}

    From the Sherman--Morrison formula, for ${t \geq 1}$,
    \begin{align}
        (\sigma_i^t)^2 &= \vec{e}_i^\top \left( \V_{t-1} + \vec{e}_{\pi_t} \vec{e}_{\pi_t}^\top / \gamma \right)^{-1} \vec{e}_i \nonumber \\
        &= \vec{e}_i^\top \left( \V_{t-1}^{-1} - \frac{\V_{t-1}^{-1} \vec{e}_{\pi_t} \vec{e}_{\pi_t}^\top \V_{t-1}^{-1}}{\gamma + \vec{e}_{\pi_t} \V_{t-1}^{-1} \vec{e}_{\pi_t}} \right) \vec{e}_i \nonumber \\
        &= (\sigma_i^{t-1})^2 - \frac{\left(\vec{e}_i^\top \V_{t-1}^{-1} \vec{e}_{\pi_t} \right)^2}{\gamma + (\sigma_{\pi_t}^{t-1})^2},
    \end{align}
    so $\sigma_i^t$ is decreasing in $t$. When $\pi_t=i$, the update depends only on the previous value $\sigma_i^{t-1}$. Consider the setting where $\pi_t=i\; \forall\; t \geq 1$.
    Then $(\sigma_i^t)^2 = {\gamma (\sigma_i^0)^2 / (\gamma + t(\sigma_i^0)^2)}$, which
    can be shown by induction. It clearly holds for $t=0$. For $t \geq 1$,
    \begin{align}
        (\sigma_i^t)^2 &= (\sigma_i^{t-1})^2 \left( 1 - \frac{(\sigma_i^{t-1})^2}{\gamma + (\sigma_i^{t-1})^2} \right) \nonumber \\
        &= \frac{\gamma (\sigma_i^{t-1})^2}{\gamma + (\sigma_i^{t-1})^2} \nonumber \\
        &= \frac{\gamma^2 (\sigma_i^0)^2}{(\gamma + (t-1)(\sigma_i^0)^2) \left(\gamma + \frac{ \gamma (\sigma_i^0)^2}{\gamma + (t-1)(\sigma_i^0)^2}\right)} \nonumber \\
        &= \frac{\gamma (\sigma_i^0)^2}{\gamma + t(\sigma_i^0)^2}.
    \end{align}
    In the setting where we do not have $\pi_t = i$ for all $t \geq 1$, since $\sigma_i^t$ is decreasing even when $\pi_t \neq i$, we can upper bound $\sigma_i^t$ with what its value
    would be if at each time $t$ such that $\pi_t \neq i$ we do not update $\sigma_i^t$. This
    would mean that by time $t$, $\sigma_i^t$ has been updated $n_i^t$ times, yielding the stated bound.

\subsection{Proof of Lemma~\ref{lemma:log-det}}

This lemma is derived from Lemma 6 of \citet{valko2014spectral}.
    If $\vec{Q} \vec{\Lambda} \vec{Q}^\top$ is the eigendecomposition of $\L_\lambda$, then let $\V_T$ and $\vec{\Lambda}$ in the notation of \citet{valko2014spectral} be equal to $\gamma \vec{Q}^\top \V_T \vec{Q}$ and $\gamma \vec{\Lambda}$, respectively, in our notation. The result follows by the invariance of determinants under unitary transformations.

\end{document}